\documentclass[conference]{ieeeconf}
\pdfoutput=1 
\IEEEoverridecommandlockouts
\usepackage{amsmath,amsfonts}
\usepackage{algorithmic}
\usepackage{algorithm}
\usepackage{array}
\usepackage{subfig}
\usepackage{textcomp}
\usepackage{stfloats}
\usepackage{url}
\usepackage{verbatim}
\usepackage{graphicx}
\usepackage{cite}
\usepackage{xcolor}
\usepackage{wrapfig}
\usepackage{hyperref}
\usepackage{orcidlink}  
\hypersetup{
    colorlinks=true,       %
    linkcolor=blue,        %
    citecolor=green,       %
    urlcolor=red           %
}
\begin{document}
\title{TwinTac: A Wide-Range, Highly Sensitive Tactile Sensor with Real-to-Sim Digital Twin Sensor Model}

\author{Xiyan Huang\orcidlink{0009-0007-5469-8298}, Zhe Xu\orcidlink{0009-0004-1043-4248} and Chenxi Xiao$^*$\orcidlink{0000-0002-7819-9633}
\thanks{This work was supported by Shanghai Frontiers Science Center of Human-centered Artificial Intelligence (ShangHAI), MoE Key Laboratory of Intelligent Perception and Human-Machine Collaboration (KLIP-HuMaCo). The experiments of this work were supported by the Core Facility Platform of Computer Science and Communication, SIST, ShanghaiTech University.}
\thanks{Xiyan Huang is with the School of Information Science and Technology at ShanghaiTech University,  {\tt\footnotesize huangxy22023@shanghaitech.edu.cn}}
\thanks{Zhe Xu is with the School of Information Science and Technology at ShanghaiTech University,  {\tt\footnotesize xuzhe2023@shanghaitech.edu.cn}}%
\thanks{Chenxi Xiao ($^*$ corresponding author) is with the School of Information Science and Technology at ShanghaiTech University,  {\tt\footnotesize xiaochx@shanghaitech.edu.cn}}%
}

\maketitle

\begin{abstract}
Robot skill acquisition processes driven by reinforcement learning often rely on simulations to efficiently generate large-scale interaction data. However, the absence of simulation models for tactile sensors has hindered the use of tactile sensing in such skill learning processes, limiting the development of effective policies driven by tactile perception. To bridge this gap, we present TwinTac, a system that combines the design of a physical tactile sensor with its digital twin model. Our hardware sensor is designed for high sensitivity and a wide measurement range, enabling high quality sensing data essential for object interaction tasks. Building upon the hardware sensor, we develop the digital twin model using a real-to-sim approach. This involves collecting synchronized cross-domain data, including finite element method results and the physical sensor's outputs, and then training neural networks to map simulated data to real sensor responses. Through experimental evaluation, we characterized the sensitivity of the physical sensor and demonstrated the consistency of the digital twin in replicating the physical sensor’s output. Furthermore, by conducting an object classification task, we showed that simulation data generated by our digital twin sensor can effectively augment real-world data, leading to improved accuracy. These results highlight TwinTac's potential to bridge the gap in cross-domain learning tasks.
\end{abstract}

\section{Introduction}\label{sec:intro}

Tactile sensing is crucial for human dexterity, enabling contact perception and reactive sensorimotor control during various manipulation tasks \cite{wei2024human}. This capability facilitates the acquisition and execution of diverse skills, such as tool use, object grasping, and assembly \cite{li2014learning, calandra2018more, hogan2018tactile, ganguly2020graspingindark}. Inspired by this biological function, robotics research has increasingly focused on integrating tactile sensing to develop autonomous interaction skills, gradually narrowing the dexterity gap between robots and humans \cite{wang2019tactile, Zhao2023GelSightSH, Romero2024EyeSightHD, lee2024dextouch, jin2023progress}. In essence, by providing robots with a mechanism to dynamically perceive and interact with their environment, tactile sensing bridges the gap between passive observation and active physical engagement.

Modern robotic systems increasingly rely on data-driven methods to interpret tactile information, which typically require large amounts of data for training \cite{schmitz2014tactile, vulin2021improved, yin2024learning}. Simulation platforms have emerged as critical tools in this context, providing cost-effective and risk-free environments for generating interaction data. Advances in soft-body contact simulation techniques (e.g., Material Point Method (MPM) and Finite Element Method (FEM)) have enabled the development of digital tactile sensor models, offering high-fidelity simulated tactile data for training manipulation tasks \cite{si2024difftactile, si2022taxim}. Along these lines, state-of-the-art work has developed simulated visuotactile sensors that combine optical rendering with deformation simulation \cite{ding2021sim}. However, reliable simulation models for non-optical tactile sensors remain absent in existing simulation frameworks. Non-optical tactile sensors are widely adopted for their simplicity, low computational overhead, and ease of integration \cite{zou2017novel}. The lack of accurate models for these sensors limits their use in simulation-driven pipelines, hindering the scalable development of tactile skills.

\begin{figure}[t]
    \centering
    \includegraphics[width=1.0\linewidth]{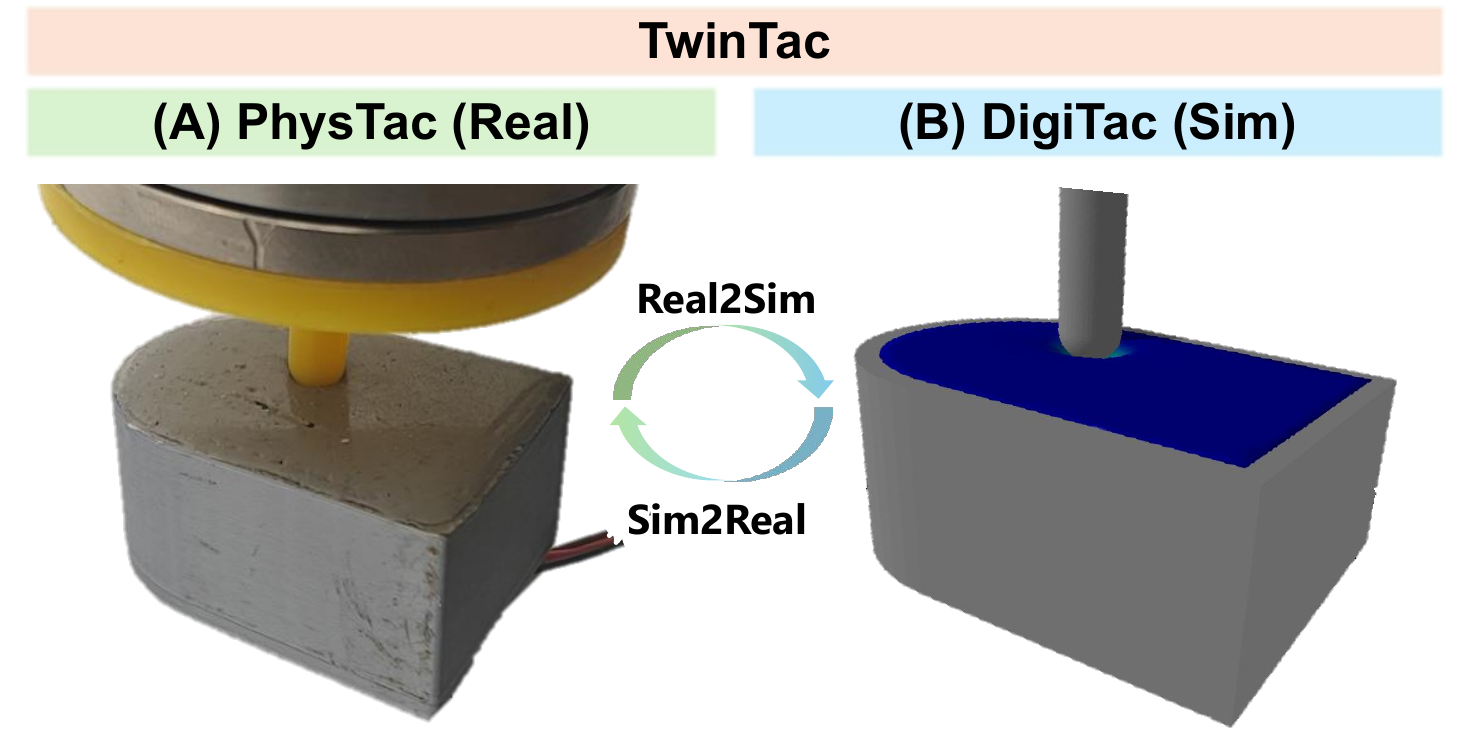}       
    \caption{We introduce TwinTac, a unified tactile system consisting of: (A) PhysTac, a physical tactile sensor with high sensitivity and a large measurement range, and (B) DigiTac, the digital twin of PhysTac.}
    \label{fig:teaser}
\end{figure}

To bridge this gap, we introduce TwinTac, a unified tactile system consisting of a physical sensor (PhysTac) and its digital twin model (DigiTac), as shown in Fig.~\ref{fig:teaser}. The PhysTac hardware integrates a matrix of MEMS pressure-sensing units within a deformable gel layer, achieving millinewton-level resolution and a force-sensing range of up to tens of kilograms. Complementing the hardware, the DigiTac sensor was developed using FEM-based simulations within the NVIDIA Isaac Gym framework \cite{nvidia2021isaacgym}. To bridge the sim-to-real domain gap, a deep neural network translates FEM simulation outputs into the PhysTac sensor’s response profiles, facilitating the development of DigiTac through Real2Sim alignment with PhysTac. Through experimental validation, we demonstrated DigiTac’s high fidelity in replicating PhysTac’s behavior. Furthermore, we showed DigiTac’s utility in data generation, assisting data-driven network training through Sim2Real cross-domain learning. These results highlight TwinTac’s ability to unify the physical and simulated domains, enabling scalable and data-efficient learning for tactile-driven robotic tasks.

The key contributions of this work are as follows: 
\begin{enumerate} \item 
\textbf{PhysTac}: The design of a tactile sensor  with high sensitivity, a wide pressure measurement range, and ease of fabrication. 
\item \textbf{DigiTac}: A high-fidelity digital twin model of the PhysTac sensor developed through Real2Sim technique. 
\item \textbf{TwinTac Evaluations}: Comprehensive experiments characterizing both the PhysTac and DigiTac sensors, along with demonstrations of how DigiTac enhances cross-domain classification tasks. 
\end{enumerate}

\section{Related Work}\label{sec:rela}

\subsection{Tactile Sensors in Robotics}
Tactile sensing is critical for enabling robots to interact with unstructured environments, providing essential feedback for manipulation, exploration, and safety \cite{dahiya2009tactile}. These tasks commonly rely on sensors based on various principles, including piezoelectric, piezoresistive, capacitive, magnetic, thermal modalities, and optics \cite{dahiya2013robotic, yamaguchi2019recent}. Among these, visuotactile sensors \cite{yuan2017gelsight, lambeta2020digit} have recently gained increased usage due to their high-quality data and compatibility with learning-based pipelines, although they have limitations, including computational overhead, elastomer's durability, and bulky form factors that restrict integration as thin surfaces \cite{zou2017novel, abad2020visuotactile}. In contrast, pressure-based sensors, such as piezoresistive (e.g., FSR arrays \cite{li2012combined}) and capacitive tactile skins, compensate for these drawbacks by offering advantages in scalability, cost-effectiveness, and low computational requirements \cite{yu2021recent}. Advances in microelectromechanical systems (MEMS) fabrication have further enhanced their utility, enabling miniaturized, high-sensitivity designs that make them viable for delicate contact tasks. Previous work has also explored the utilization of MEMS for designing tactile sensors \cite{tenzer2014feel, kim2022barotac, koiva2020barometer}. However, previous designs have barely achieved a wide force-sensing range exceeding that of the human finger (approximately 100 N \cite{fingerforce}) while maintaining ultra-high sensitivity. In this paper, we propose a new MEMS-based tactile sensor designed to bridge this gap. A comparison with previous MEMS-based sensors is provided in Table~\ref{tab:sensor_comparison}. We highlight that TwinTac offers a significantly higher dynamic range, along with a simulation framework that shows promise for large-scale data collection.

\subsection{Simulation of Tactile Sensors}
Tactile sensor simulation offers a cost-effective and efficient means of acquiring tactile-based skills, serving as a valuable data source for learning-based approaches \cite{akinola2024tacsl}. Recent advancements have primarily focused on developing simulations for visuotactile sensors, such as GelSight~\cite{yuan2017gelsight} and Digit~\cite{lambeta2020digit}, which excel at rendering high-resolution images that capture detailed local geometric information about objects \cite{li2024vision}. Early simulation methods for these sensors relied on rigid-body geometry \cite{si2022taxim, xu2023efficient}, but more recent approaches have incorporated soft-body dynamics using techniques like FEM \cite{si2024difftactile, peng20243d} and MPM \cite{zhang2023tirgel}. More recent solvers, such as incremental potential contact \cite{du2024tacipc}, further balance simulation fidelity and speed, making the simulation of tactile sensors more practical and usable.

In contrast, the simulation of non-visuotactile sensors remains largely unexplored. While rigid-body physics engines can simulate basic force and torque sensors, high-fidelity models for broader real-world devices are still barely developed. To date, only the BioTac sensor has been successfully modeled using data-driven techniques \cite{narang2021sim}, leaving many widely used tactile sensors without accurate simulators. To bridge this gap, we propose a hybrid simulation framework that integrates FEM-based deformation modeling with a data-driven network, offering a high fidelity solution to faithfully replicate the behavior of our proposed PhysTac tactile sensor.

\begin{table}[t]
\vspace{2mm}
\centering
\caption{Comparison with previous MEMS tactile sensors.}
\label{tab:sensor_comparison}
\resizebox{\columnwidth}{!}{
\begin{tabular}{cccc}
\hline\hline
\textbf{Device} & \textbf{MEMS Model} & \textbf{Min Force (N)} & \textbf{Max Force (N)} \\ \hline
PhysTac (Ours) & CPS135B-1500D & \textbf{0.01} & \textbf{\textgreater200} \\ 
Takkitile \cite{tenzer2014feel} & MPL115A & 0.01 & 5 \\ 
BaroTac \cite{kim2022barotac} & BMP384 & 0.015 & 9 \\ 
SDRH Palm \cite{koiva2020barometer} & BMP388 & NR & 3 \\ \hline\hline
\end{tabular}
}
\vspace{-4mm}
\end{table}

\begin{figure*}[t]
    \centering
    \includegraphics[trim=10pt 65pt 5pt 5pt, clip, width=0.95\textwidth]{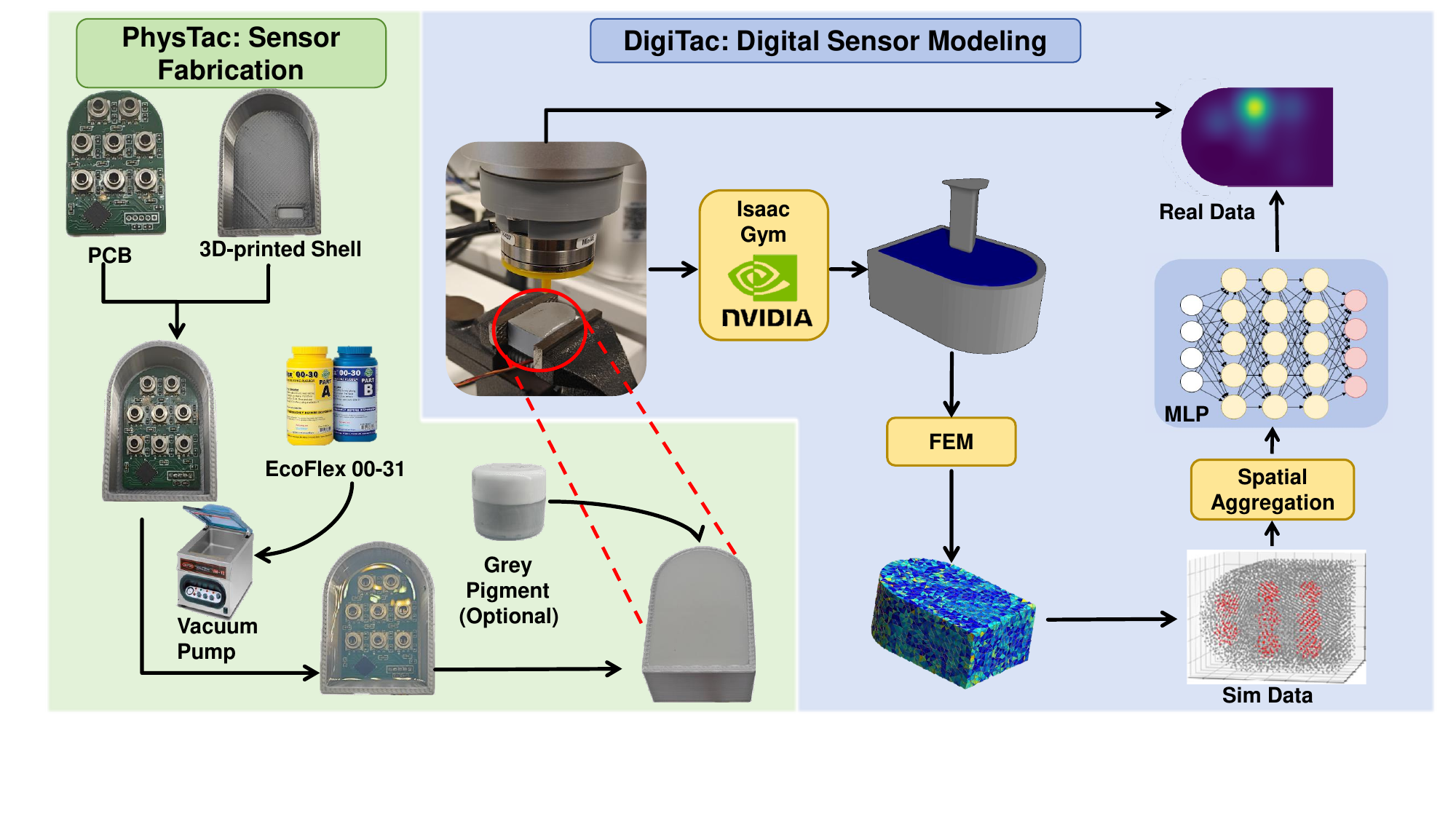}  
    \caption{Technical pipeline of TwinTac, including the procedures for fabricating the PhysTac sensor through gel casting, as well as the techniques used for data collection and the creation of the DigiTac sensor via Real2Sim.}
    \vspace{-4mm}
    \label{fig:pipeline}
\end{figure*}

\section{Method}

This section outlines the design of the TwinTac system. The overall framework of TwinTac is illustrated in Fig.~\ref{fig:pipeline}. Our methodology consists of two main components. First, we introduce the PhysTac hardware, along with the techniques for extracting spatial pressure distribution from taxels (Sec.~\ref{sec:sensor_fab}). Second, we elaborate on the data collection protocol (Sec.~\ref{sec:met:real2sim}), and the procedure for developing the DigiTac digital twin model (Sec.~\ref{sec:prediction}), respectively.

\subsection{PhysTac Sensor} \label{sec:sensor_fab}
First, we outline the procedures for fabricating the PhysTac sensor. Our sensor system consists of three main components: 1) a 3D-printed sensor shell, 2) a printed circuit board (PCB) with pressure sensors, and 3) an elastomer interface designed to interact with the objects.

\textbf{Printed Circuit Board.} The key component of PhysTac is the taxel matrix on the printed circuit board, which consists of eight MEMS pressure-sensing taxels. These taxels are CPS135B-1500D (Consensic Inc.), soldered onto the top layer of the circuit board, with each component featuring an I2C interface. The I2C lines of all components are multiplexed via a TCA9548A chip, simplifying communication into a single bus and achieved a data rate of 55Hz for all taxel channels. Although previous works have explored using similar barometer sensors to create tactile sensors, our design achieves a significantly wider pressure measurement range (refer to Table~\ref{tab:sensor_comparison}). The CPS135B-1500D offers a resolution of 0.01 kPa, comparable to the sensitivity of human skin \cite{johansson1979tactile}, and a detection range of 50 to 1500 kPa (15× atmospheric pressure), which exceeds the limit of puncture pressure that human skin can tolerate (approximately 700 kPa \cite{dailiana2008injection}).

\textbf{Elastomer.} The sensor elastomer is fabricated using Ecoflex 00-31 silicone gel (Smooth-On Inc.). This gel is selected for its low cost, adequate elasticity for deformation recovery, and durability under large forces. The gel-casting process involves the following steps. First, a shell is 3D-printed, with the sensor board fixed inside. Second, silicone gel is poured into the assembly. The gel is prepared by mixing parts A and B in a 1:1 ratio, stirring for three minutes, and curing at room temperature for four hours. Pigments can be added to the Ecoflex to achieve the desired coloration. To ensure homogeneity, air bubbles are removed from the gel mixture using a vacuum pump. Fabrication details are illustrated in Fig.~\ref{fig:pipeline}.

\textbf{Pressure Visualization}.  To visualize the spatial distribution of pressure from the MEMS sensor matrix, we employ Radial Basis Function (RBF) interpolation to construct a continuous pressure map. The RBF interpolation estimates the pressure at any location $\mathbf{x}$ using the following equation:
\begin{equation}
    f(\mathbf{x}) = \sum_{i=1}^{N} (\lambda_i - p(\mathbf{x}_i)) \varphi(\|\mathbf{x} - \mathbf{x}_i\|)
\end{equation}
where $i$ is the taxel's index,  $f(\mathbf{x})$  is the interpolated pressure; $\lambda_i$ is an offset pressure value at the sensor location $\mathbf{x}_i$ (used to ensure zero baseline when without external load); $p(\mathbf{x}_i)$ is the measured pressure at that sensor location; $\varphi(\|\mathbf{x} - \mathbf{x}_i\|)$ is the radial basis function (Gaussian), calculated using the Euclidean norm as distance. A pressure distribution matrix can be generated for visualization by iterating this calculation over mesh grid points on the sensor surface.

\subsection{Real2Sim Data Collection}\label{sec:met:real2sim}

Next, we outline the procedures for data collection to support the development of the DigiTac model.

\textbf{Simulation Setup.} First, the DigiTac sensor's simulation was set up within Isaac Gym~\cite{nvidia2021isaacgym}. The sensor is 3D-modeled as two CAD components: an elastomer (soft body) and a rigid shell. The elastomer is simulated using Isaac Gym’s flexible material engine. To prepare for simulating elastomer's deformation, the elastomer's volumetric meshes are generated using fTetWild~\cite{10.1145/3386569.3392385}. To balance computational efficiency and simulation accuracy, the elastomer is modeled using 7636 vertices (38599 tetrahedrons). %

In addition to the elastomer, the rigid shell constrains the deformation of the elastomer by encapsulating its bottom and side surfaces, leaving only the top surface exposed for interaction with external objects. One challenge we encountered was handling large deformations in FEM simulations. To ensure numerical stability, we introduced a 1 mm gap between the soft body and the side shell, providing additional space for expansion.

\begin{figure}[t]
    \centering
    \includegraphics[width=0.9\linewidth]{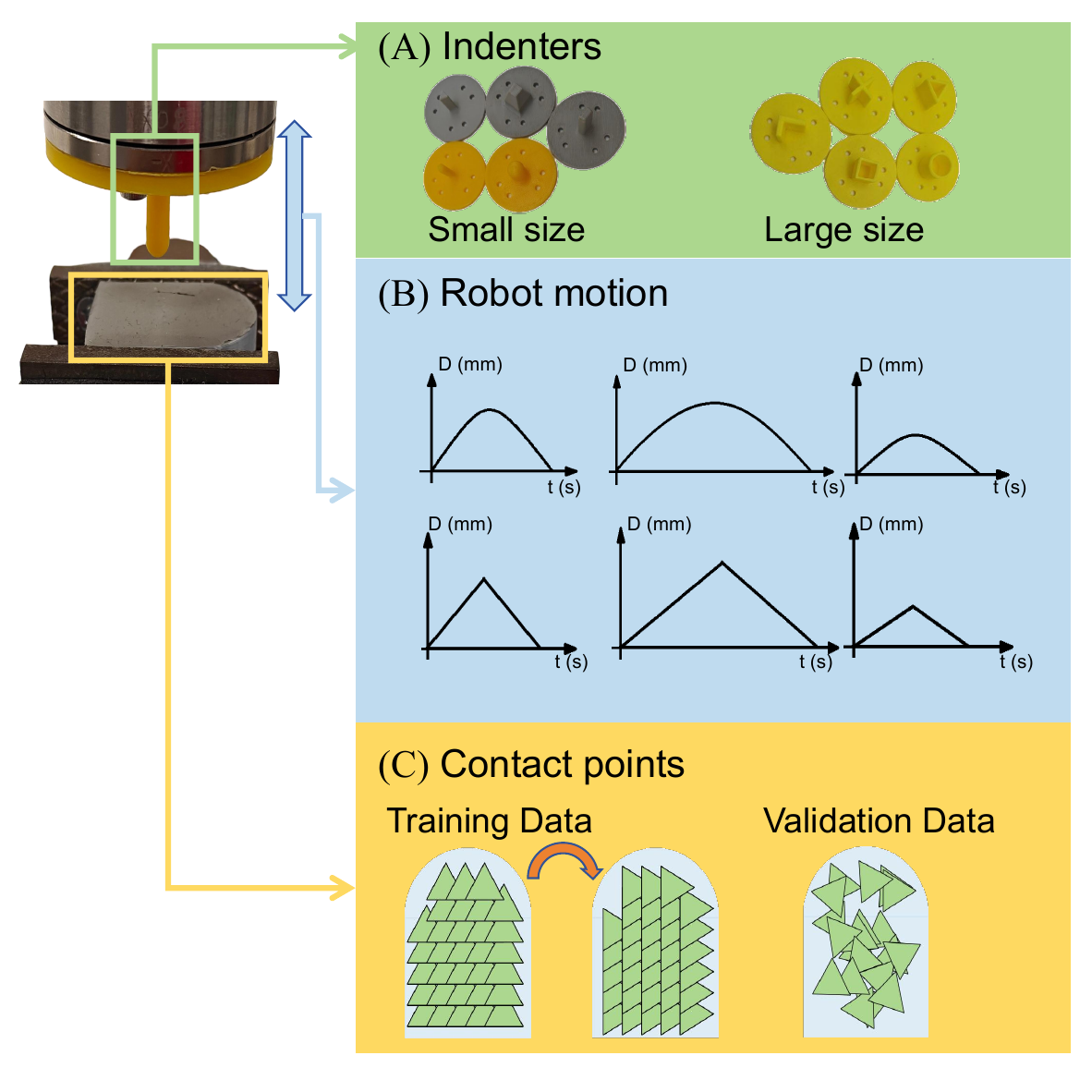}       \caption{Data collection protocol for creating DigiTac, including (B) indenters used, (A) robot action sequences for indentation depth $D$, and (C) dataset construction.}
    \label{fig:data_collection}
\end{figure}

\textbf{Real Data Collection Setup.} A real-world data collection platform was set up to facilitate Real2Sim. To achieve this, we mounted an ATI Mini45 FT sensor on the end effector of the X-Arm 6 robot. Then, we mount an indenter (3D printed, as shown in Fig.~\ref{fig:data_collection}) onto FT sensor. The whole arm is controlled using the $\text{set\_servo\_cartesian}$ command.

\textbf{Data Collection.} Data are then collected by applying the same indentation conditions in both the simulation and real-world platforms. For simulation data collection, two different types of information are collected, which are denoted as: $\mathcal{X}$ (refer to Eq.(\ref{eq:Sim_X})) and $\mathcal{F}_{\mathrm{sim}}$ (refer to Eq.(\ref{eq:Sim_F})).

\begin{equation}
\label{eq:Sim_X}
    \mathcal{X} = \{\mathbf{X}_t\}_{t=1}^T,\quad 
    \mathbf{X}_t = \begin{bmatrix}
    x_1^{(t)} & y_1^{(t)} & z_1^{(t)} & s_1^{(t)} \\
    \vdots    & \vdots    & \vdots    & \vdots    \\
    x_n^{(t)} & y_n^{(t)} & z_n^{(t)} & s_n^{(t)}
    \end{bmatrix} \in \mathbb{R}^{n \times 4}
\end{equation}

Where $x_i^{(t)},y_i^{(t)},z_i^{(t)}$ denote the centroid coordinates of the $i$-th tetrahedron  node at time step $t$, and $s_i^{(t)}$ represents node's Von Mises stress~\cite{vonmises1913}. Besides, we also collect $\mathcal{F}_{\mathrm{sim}}$, where $F_{\mathrm{z}}^{(t)}$ represents the axial force exerted by simulated indenter.

\begin{equation}
\label{eq:Sim_F}
    \mathcal{F}_{\mathrm{sim}} = \{^{sim}F^{(t)}_{\mathrm{z}}\}_{t=1}^T,\quad 
    ^{sim}F^{(t)}_{\mathrm{z}} \in \mathbb{R}
\end{equation}

In physical environment, we record the pressure readings of the MEMS sensor (denoted as $\mathcal{S}$, refer to Eq.~(\ref{eq:Signal_real})). We also record the FT sensor's force $\mathcal{F}_{\mathrm{real}}$ (Eq.~(\ref{eq:F_real})), for the purpose of timing synchronization and further ablation studies.

\begin{equation}
\label{eq:Signal_real}
    \mathcal{S} = \{\mathbf{s}_t\}_{t=1}^T,\quad 
    \mathbf{s}_t \in \mathbb{R}^8
\end{equation}

\begin{equation}
\label{eq:F_real}
    \mathcal{F}_{\mathrm{real}} = \{^{real}F_{z}^{(t)}\}_{t=1}^T,\quad 
    ^{real}F_{z}^{(t)}\in \mathbb{R}
\end{equation}

Where $\mathbf{s}_t$ captures the tactile pressure distribution from the 8-channel pressure sensing units, and $^{real}F_{z}^{(t)}$ represents the axial contact force measured by the ATI FT sensor. To maximize dataset diversity, we adopted different indenters and varied both the indentation locations and the sampling action sequence (refer to Fig.~\ref{fig:data_collection} (C)).

\textbf{Data Preprocessing.} Due to unavoidable robot positioning errors and hardware delays, the real and simulated data may be misaligned in time. Therefore, an alignment was computed between $\mathcal{F}_{\mathrm{real}}$ and $\mathcal{F}_{\mathrm{sim}}$ using Dynamic Time Warping (DTW)~\cite{muller2007dynamic}. Once mapping relationship is obtained, we wrap the pressure sensor's readings $\mathcal{S}$ using DTW results.

\subsection{Training DigiTac Sensor Model}
\label{sec:prediction}
\begin{figure}[t]
    \centering
    \includegraphics[width=0.9\linewidth]{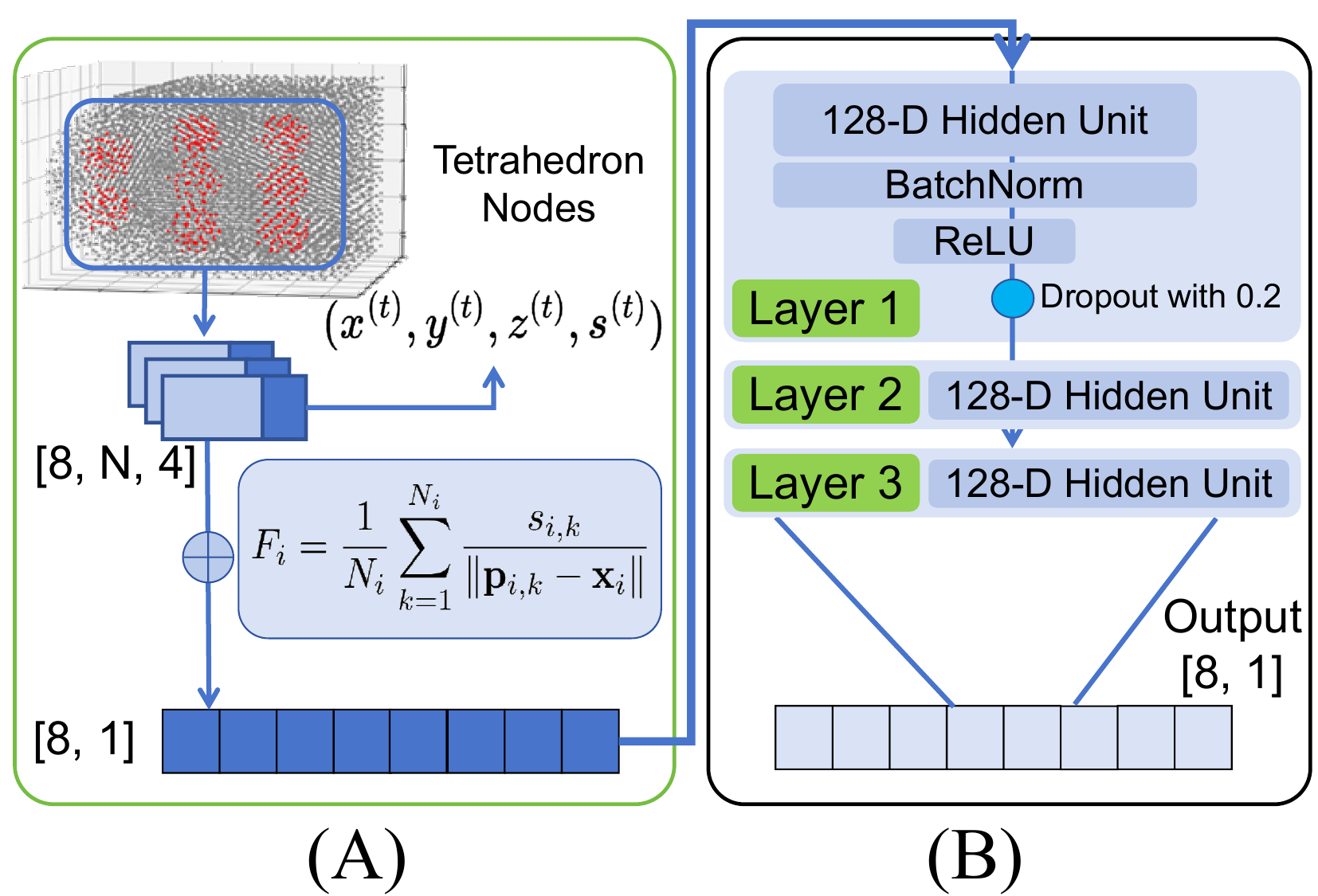}
    \caption{DigiTac's network design.
    (A) Aggregated stress features from neighboring tetrahedron nodes, followed by (B) MLP network that infers the sensor's signal outputs.}
    \label{fig:network}
    \vspace{-2mm}
\end{figure}

While the stress-strain data of the elastomer can be obtained from the simulation, these values do not directly reflect sensor's output signals. Therefore, DigiTac's sensor model is created through a mapping from the FEM output to the sensor signals: $ \Phi: \mathcal{X} \mapsto \mathcal{S}$. This process can be achieved efficiently using a lightweight network structure, as shown in Fig.~\ref{fig:network}.

Due to the large number of FEM nodes, processing the information from the entire elastomer body is computationally expensive. In practice, we found that extracting a predefined subset of nodes could yield satisfactory performance. The nodes used for this subset are located on the sensor's surface and directly above the 8 positions of taxels (red nodes in Fig.~\ref{fig:network} (A), which includes only 609 nodes. To further reduce dimensionality, we found that aggregating the stress from all nodes (Eq.~(\ref{eq:avg_stress})) could provide informative features for the sensor output, which was used as network's input.

\begin{equation}
\label{eq:avg_stress}
    F_i = \frac{1}{N_i} \sum_{k=1}^{N_i} \frac{s_{i,k}}{\|\mathbf{p}_{i,k} - \mathbf{x}_i\|} \quad (i = 1,\dots,8)
\end{equation}
where $\mathbf{x}_i \in \mathbb{R}^3$ denotes the location of the $i$-th sensor unit, $\mathbf{p}_{i,k} = (x_{i,k}, y_{i,k}, z_{i,k})^\top$ represents 3D coordinates of the $k$-th node point around the $i$-th sensor unit, and $s_{i,k} \in \mathbb{R}$ represents Von Mises stress at $\mathbf{p}_{i,k}$. $N_i$ indicates the number of points within $i$-th cluster.

The network was trained on 36k paired temporal frames using L1 loss. For domain randomization, we injected zero-mean Gaussian noise ($\sigma$=0.05). Optimization was performed using the AdamW optimizer with gradient clipping (max norm = 1.0), and adaptive learning rate reduction (initial learning rate = 1e-3, scaling factor = 0.3 upon validation plateau).

\section{Experiments}

In this section, we assess TwinTac through a series of contact characterization experiments. First, the PhysTac sensor is evaluated by analyzing its response to external contacts (Sec.~\ref{sec:exp:PhysTac}). Second, the DigiTac sensor is then evaluated by comparing its simulated output with the data from the PhysTac sensor. Finally, the DigiTac sensor is tested for its potential to assist in data-driven tasks via Sim2Real.

\subsection{Characterization of PhysTac Sensor} \label{sec:exp:PhysTac}

\begin{figure}[t]
    \centering
    \includegraphics[width=0.48\textwidth]{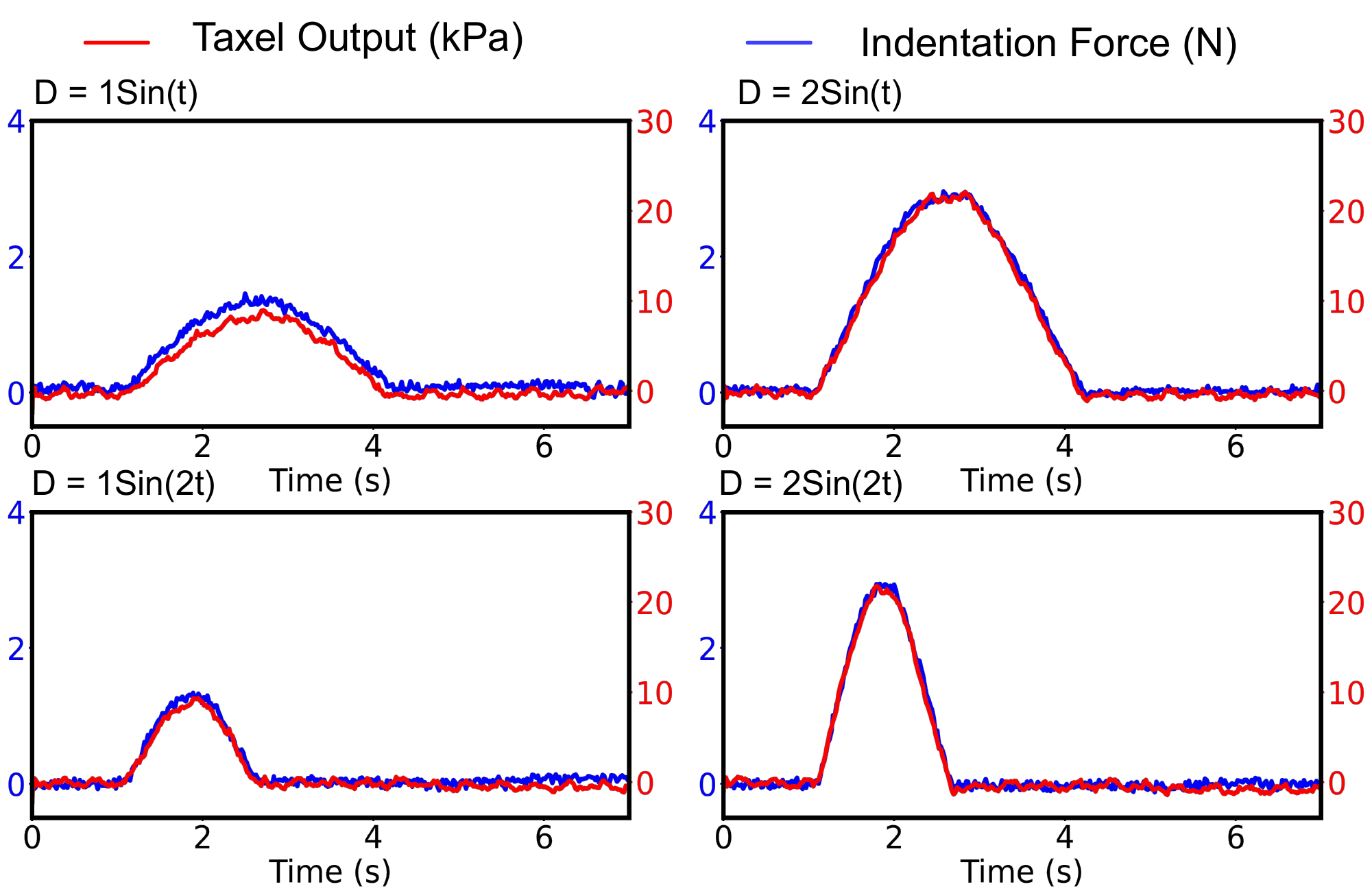}  
    \caption{Force response characterization in the time domain at varying amplitudes and frequencies. The $F_z$ is measured in Newtons, and the sensor unit's response is in kPa.}
    \label{fig:exp1-1}
    \vspace{-2mm}
\end{figure}

\begin{figure}[t]
\vspace{2mm}
    \centering
    \includegraphics[width=1.0\linewidth]{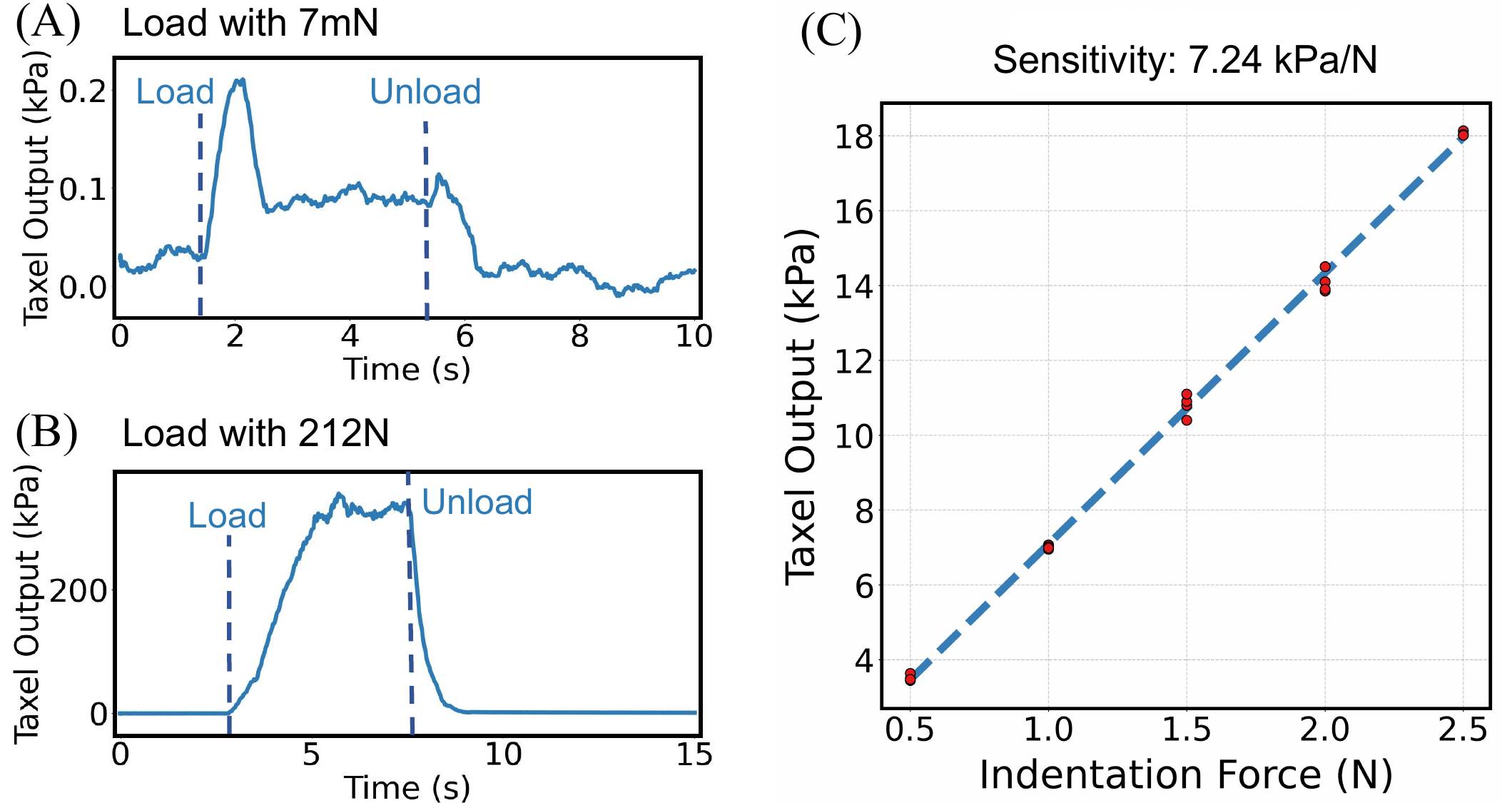}  
    \caption{Characterization of the PhysTac taxel's measurement range and sensitivity using a 25 mm round indenter.  }
    \label{fig:exp1-2}
    \vspace{-4mm}
\end{figure}

Experiments were conducted to characterize the PhysTac sensor for evaluating its data fidelity. To achieve this, we characterized each taxel's response to external forces, aiming to reveal its time-domain characteristics and sensitivity. This was accomplished through an indentation experiment using a round indenter with a 25 mm diameter. The indentation was positioned directly above the characterized taxel. The characterization results are shown in Fig.~\ref{fig:exp1-1} and Fig.~\ref{fig:exp1-2}.

Fig.~\ref{fig:exp1-1} shows the time-domain characteristics of the taxel's output. In this figure, both the taxel's output and the axial force of the indenter (measured by the FT sensor) are visualized. The results demonstrate that the taxel's output closely correlates with the indentation force data. This correlation remains consistent as the indenter varies in speed and magnitude, highlighting the taxel's high fidelity in characterizing external forces.

Fig.~\ref{fig:exp1-2} characterizes the taxel's measurement range in response to external contact forces, as well as its sensitivity. Specifically, Fig.~\ref{fig:exp1-2} (A) showcases the sensor's capability to detect small forces. When loading the sensor with only 0.007 N of force, we observed a step response, with the sensor's steady-state output increasing untill unloaded. Besides, the sensor is also capable of detecting large contact forces (212 N of force), as shown in the experiment in Fig.~\ref{fig:exp1-2} (B). Furthermore, we characterize the sensor's sensitivity in Fig.~\ref{fig:exp1-2} (C). We observed a linear relationship between the sensor's output and the applied external force, with a sensitivity of approximately 7.24 kPa/N.

\begin{figure}[t]
    \centering
    \includegraphics[width=1.0\linewidth]{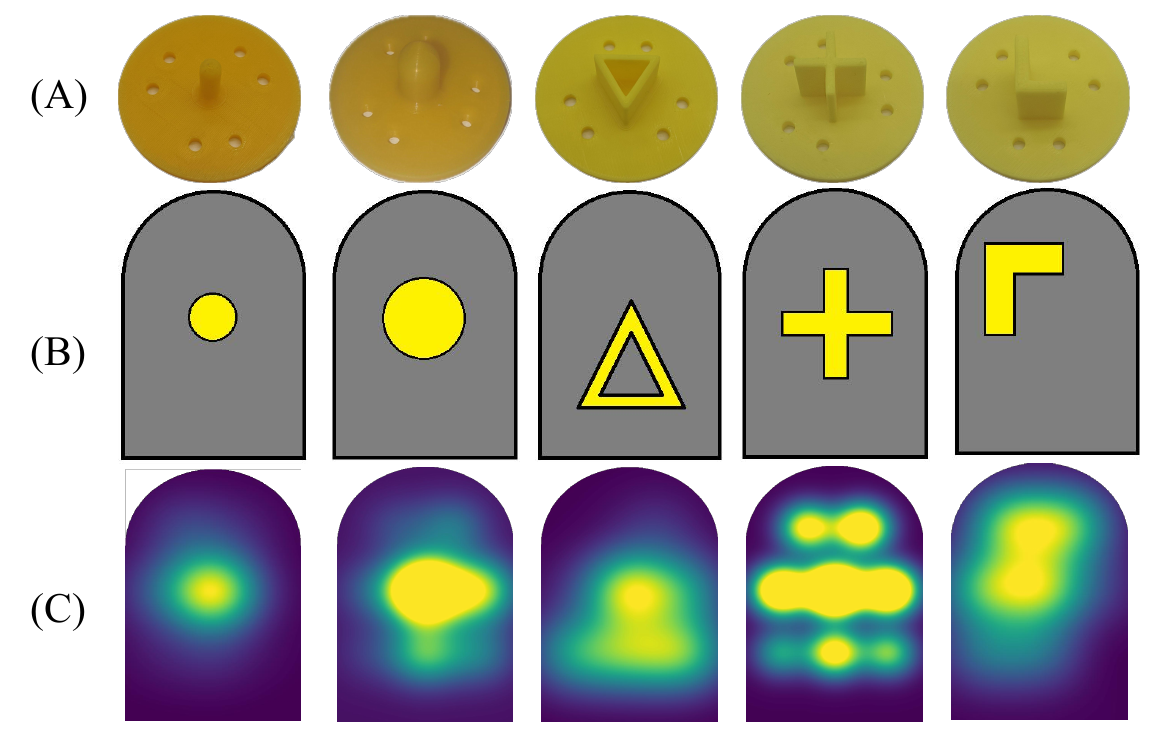}  
     \caption{Characterization of PhysTac using different indenters. (A) Shape of Indenters. (B) Locations where indentation applied. (C) Sensor's reading (with RBF interpolation). }
    \label{fig:exp1-3}
    \vspace{-3mm}
\end{figure}

Beyond individual taxels, we also evaluated the sensor's performance in characterizing spatial pressure distribution. To validate this, we indented the sensor with five different indenters of varying shapes, as shown in Fig.~\ref{fig:exp1-3}. The sensor's output data were then collected for visualization using the technique introduced in Sec.~\ref{sec:sensor_fab}. Qualitatively, the visualization results reflect the shape of the indenter. Therefore, we conclude that the sensor is capable of providing a spatial distribution of contact pressure, enabling the discrimination of object shape. Nevertheless, the visualization is generated using only 8 MEMS taxels, which provides insufficient resolution to reveal surface textures and subtle geometric differences.

\subsection{Characterization of DigiTac Sensor Model}\label{sec:exp:DigiTac}
\begin{figure}[htbp]
    \centering
    \includegraphics[width=1.0\linewidth]{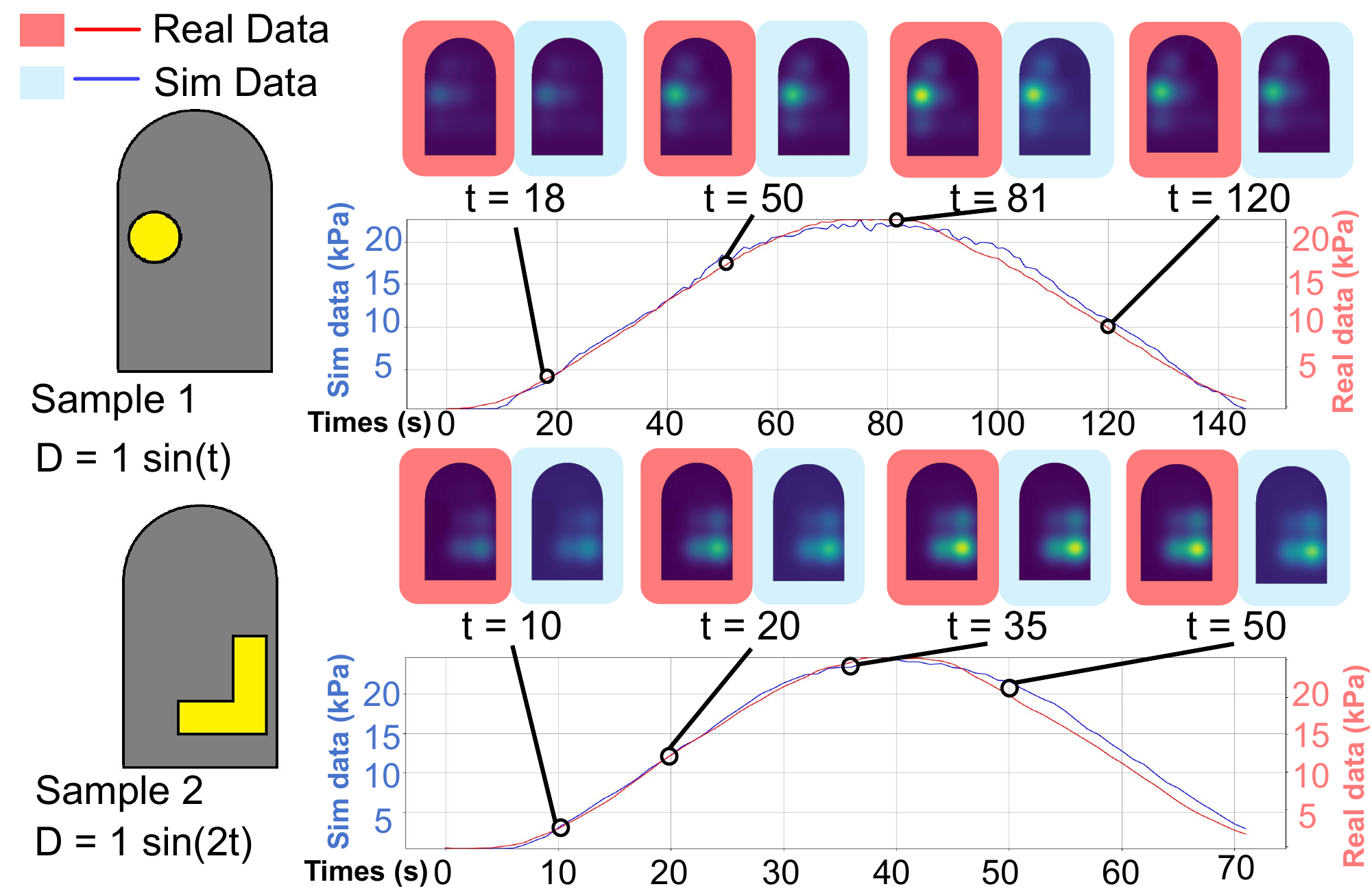}  
    \caption{Comparison of DigiTac's output versus PhysTac's output under the same indentation condition.}
    \label{fig:exp2}
    \vspace{-4mm}
\end{figure}

Next, we assess the fidelity of the DigiTac model by comparing its output with the ground truth data from the PhysTac sensor. The comparison results are shown in Fig.~\ref{fig:exp2}. Here, we showcase two indentation samples. It can be observed that the DigiTac model's predictions match the ground-truth data from PhysTac.  On average, the prediction error corresponds to 5.95\% of the maximum contact pressure. This demonstrates that the DigiTac model can replicate the PhysTac's output with high fidelity. %

\subsection{DigiTac Sensor for Sim2Real}\label{sec:exp:Sim2Real}

Lastly, experiments were conducted to showcase the DigiTac's capability in producing high-fidelity data that can offload the demand for real data in data-driven tasks. This demonstrates the practical value of the DigiTac model, as data collection in the real world is time-consuming and intrusive to the environment due to the contact forces involved.

\begin{figure}[ht]
    \centering
    \includegraphics[width=1.0\linewidth]{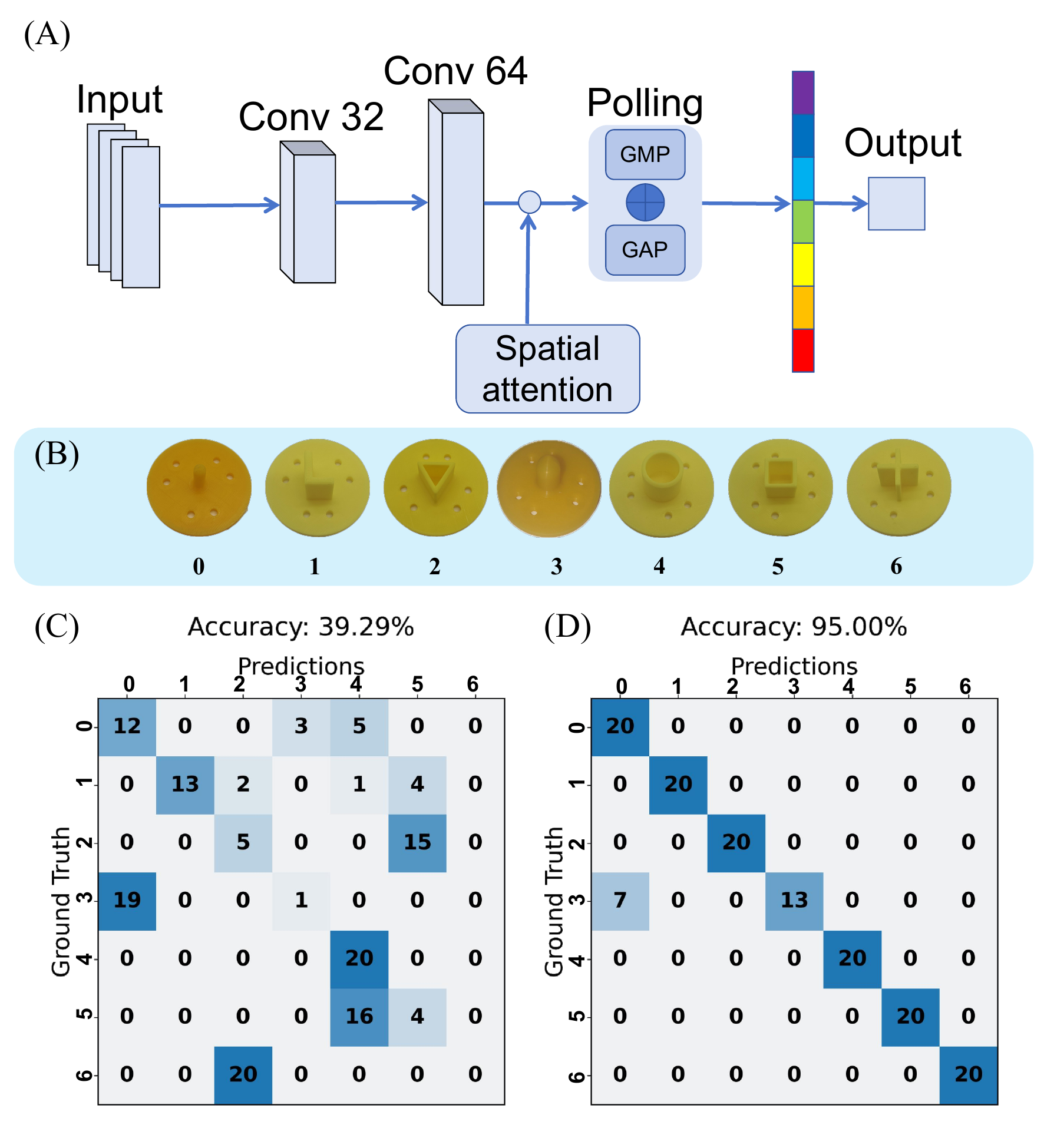}  
    \caption{Network structure and confusion matrices for the classification task. (A) Time-series transform network. (B) Indenters used for object classification. (C) Confusion matrix obtained by training solely on real-world data (test accuracy: 39.29\%). (D) Confusion matrix obtained by training on augmented data generated via DigiTac (test accuracy: 95\%).}
    \label{fig:cm}
    \vspace{-2mm}
\end{figure}

To leverage the DigiTac sensor for data augmentation, we set up an indentation experiment in the simulation for data collection. The indentation was performed with varying depths, speeds, and rotational angles to enhance the dataset. At the same time, a real-world indentation experiment was conducted to collect a correlated data set, although with much smaller quantities. A classification neural network was then designed to classify the real-world indentation data using a time-series Transformer \cite{wen2022transformers} (Fig.~\ref{fig:cm} (A)). This network accepts the sensor’s time-series data as input and classifies the shape of the indenters.

Comparison experiments were conducted with two settings: 1) Training with real-world data only: For each indenter, the dataset has 20 trials of real-world data for training, with the remaining 20 trials used for testing. 2) Training with hybrid data: For each indenter, the training set included 20 real-world trials and 96 simulation trials, and the testing set includes the remaining 20 real-world trials. The second case represents a Sim2Real setting, designed to demonstrate the usability of the simulation by simplifying data collection through the use of simulated data. 7 indenters were used for classification (Fig.~\ref{fig:cm} (B)).

The confusion matrices for the two experimental cases is shown in Fig.~\ref{fig:cm} (C). The results reveal that the time series Transformer \cite{wen2022transformers} struggled to learn effectively from the limited real-world data, achieving only 33.57\% classification accuracy. The network almost completely failed to differentiate between the spherical, cubic, and X-shaped indenters. In contrast, when synthesized data was incorporated into the training, performance improved significantly, reaching 95.0\% accuracy. This demonstrates that the DigiTac sensor effectively enhances classification by providing high-fidelity data through Sim2Real.

\section{Conclusions}
Robot skill learning often requires accurate simulation of sensor responses to ensure reliable Sim2Real transfer. In this paper, we present TwinTac, a system that integrates a physical tactile sensor (PhysTac) and its digital twin model (DigiTac), which faithfully replicates the behavior of PhysTac. The PhysTac sensor is designed with high sensitivity and a wide measurement range, exceeding the performance of earlier MEMS-based tactile sensors. The corresponding simulation model was then developed using a Real2Sim approach, aligning real-world indentation data with Isaac Gym's soft-body simulation results to predict the sensor’s output signals. Experimental results demonstrate that the simulated sensor model faithfully replicates the responses of the hardware sensor under different external contact force conditions. Furthermore, we validated DigiTac’s capability to provide high-fidelity data for data augmentation, which successfully bridged the Sim2Real gap in an object shape classification task.

One limitation of our system is its spatial resolution, which is constrained by the number of MEMS taxels in the current design. Future work will focus on enhancing the sensor's spatial resolution and expanding to additional sensing modalities, such as temperature and mechanical vibration. Moreover, we plan to extend the digital twin framework to broader robotic applications, and testing its capability to support the development of tactile skills, such as grasping and in-hand manipulation through reinforcement learning techniques.

\section*{Acknowledgment}
We faithfully thank Ziyuan Tang, Yatao Leng and Xuhao Qin for gel casting and 3D printing, Tianyi Gao for assisting with programming the I2C protocol.

\clearpage
\bibliographystyle{ieeetr} %
\bibliography{reference} %
\vfill
\end{document}